\documentclass{article} % For LaTeX2e
\usepackage{iclr2020_conference,times}

\usepackage{amsmath,amsfonts,bm}

\def\eqref#1{equation~\ref{#1}}
\def\1{\bm{1}}

\DeclareMathAlphabet{\mathsfit}{\encodingdefault}{\sfdefault}{m}{sl}
\SetMathAlphabet{\mathsfit}{bold}{\encodingdefault}{\sfdefault}{bx}{n}

\usepackage{amssymb}
\usepackage{hyperref}
\usepackage{url}
\usepackage{graphicx,float}
\usepackage{subcaption}
\usepackage{xspace}
\usepackage{comment}
\usepackage{setspace}

\title{Privileged Information Dropout\\in Reinforcement Learning}

\author{Pierre-Alexandre Kamienny\thanks{Correspondence to \href{mailto:pierrealex1995@gmail.com}{pierrealex1995@gmail.com}}\\ University of Oxford\\
\And
Kai Arulkumaran\\ Imperial College London\\
\AND
Feryal Behbahani\\ Imperial College London\\
\And
Wendelin Boehmer\\ University of Oxford\\
\And
Shimon Whiteson \\ University of Oxford}

\newcommand{\pid}{PI-Dropout\xspace}

\iclrfinalcopy % Uncomment for camera-ready version, but NOT for submission.
\begin{document}

\maketitle

\begin{abstract}

Using privileged information during training can improve the sample efficiency and performance of machine learning systems. This paradigm has been applied to reinforcement learning (RL), primarily in the form of distillation or auxiliary tasks, and less commonly in the form of augmenting the inputs of agents. In this work, we investigate Privileged Information Dropout (\pid) for achieving the latter which can be applied equally to value-based and policy-based RL algorithms. Within a simple partially-observed environment, we demonstrate that \pid outperforms alternatives for leveraging privileged information, including distillation and auxiliary tasks, and can successfully utilise different types of privileged information. Finally, we analyse its effect on the learned representations.

\end{abstract}

\section{Introduction}
The idea of incorporating additional information during training to accelerate the learning process has a long history, mainly in the context of supervised learning, and has been studied under different names, including learning using hints \citep{abu1990learning}, hidden or privileged information \citep[PI;][]{vapnik2009learning,vapnik2009new}, transfer learning \citep{pan2009survey}, and multitask learning \citep{caruana1997multitask}. By leveraging the extra information during training as either extra inputs, extra outputs or additional loss terms, these methods add inductive biases to improve generalisation.

We focus on such a setting  where privileged input features, $\mathbf{x}^*$ are available during training, alongside the regular input features, $\mathbf{x}$. In particular, we consider three broad ways of leveraging PI:
\mbox{\textit{1) Distillation:}} Using a ``teacher'' model, which receives PI as input, to regularise a separate ``student'' model, which only receives regular features as input \citep{hinton2015distilling,lopez2015unifying}.
\textit{2) Auxillary tasks:} Constructing additional loss terms based on PI \citep{caruana1997promoting}.
\textit{3) Augmenting inputs:} Using PI as additional inputs.

Previous work on using PI in RL has focused largely on distillation \citep{parisotto2015actor,rusu2015policy} and auxiliary tasks \citep{mirowski2016learning}. The disadvantage of distillation is that it requires training a separate teacher model. In settings where the PI cannot be directly predicted from regular inputs, using PI instead as input would be more beneficial. For example, predicting depth from an RGB view of a scene has been shown to be a useful auxiliary task \citep{mirowski2016learning}, but predicting non-overlapping views is in general not possible. However, multi-view integration---using these non-overlapping views as input---can be beneficial for tasks such as navigation.

Using PI as input in an actor-critic setting has been explored where only the critic has access to PI during training \citep{pinto2017asymmetric}. Other work has extended this principle to multi-agent systems, where information about the full state is used to train a centralised critic with decentralised actors \citep{lowe2017multi,foerster2018counterfactual,rashid2018qmix}.

In contrast, we explore a more general way to provide PI as input to both policy and value functions through the use of \pid \citep{lambert2018deep}, which extends information dropout \citep{achille2018information} to leverage PI. Proposed in a supervised learning setting, \pid theoretically combines the representation learning advantages of information dropout with the ability to marginalise out $\mathbf{x}^*$ during execution. Our contribution is to investigate the use of \pid in an RL setting and analyse its effect on trained agents. In our experimental setup we find that \pid outperforms other methods using PI, including auxiliary prediction tasks and distillation. In a key ablation, we show that \pid outperforms standard information dropout, indicating that our method does indeed benefit from using PI. Of note is parallel research from \cite{salter2019attention}, who also introduced a method that allows PI to be used as inputs for policies and value functions. Their method relies on aligning the attention of separate actor-critic networks, where one pair has access to PI, and is orthogonal and complementary to our work.

\section{Background and Method}
\pid \citep{lambert2018deep} is motivated by information bottleneck (IB) theory \citep{tishby2000information}. In IB, for inputs $\mathbf{x}$ and task outputs $\mathbf{y}$, an ``optimal'' intermediate representation, $\mathbf{z}$, should have high mutual information (MI) with the outputs, $I(\mathbf{z}; \mathbf{y})$, while minimising task-irrelevant information from the inputs, $I(\mathbf{x}; \mathbf{z})$. The intuition is that the learned encoding $\mathbf{z}$ will ignore uninformative patterns in $\mathbf{x}$, and therefore generalize well to previously unseen samples. It is possible to directly optimise for this property by using a relaxed Lagrangian formulation of the IB. This can be expressed as the minimisation of a cross-entropy term and an MI term, $H(\mathbf{y}|\mathbf{z}) + \beta I(\mathbf{x}; \mathbf{z})$, with hyperparameter $\beta$. Minimising the MI term corresponds to minimising the KL divergence between $p(\mathbf{z}|\mathbf{x})$ and the marginal $p(\mathbf{z})$.
 
Information dropout \citep{achille2018information} uses multiplicative dropout to implement the IB Lagrangian, where the variance of the noise applied to $\mathbf{z}$ is conditioned on $\mathbf{x}$. \pid instead conditions the noise on $\mathbf{x}^*$, allowing PI to mask out irrelevant features with noise. Concretely, in \pid, \mbox{$\mathbf{z} = h(\mathbf{x}; W) \odot Lognormal(\mathbf{0}, h^*(\mathbf{x}^*; W^*))$}
, where $h(\cdot; W)$ and $h^*(\cdot; W^*)$ represent separate subnetworks\footnote{In practice we restrict the variance $\in (0, 1)$ using a sigmoid function after $h^*(\mathbf{x}^*; W^*)$.}, parameterized by weights $W$ and $W^*$, respectively. \citet{achille2018information} use a log-normal distribution for $p(\mathbf{z})$, which in \pid results in maximising the regularisation term  $\beta\log h^*(\mathbf{x}^*; W^*)$. During execution, the PI subnetwork is ignored, and the variance is manually set to 0, marginalising out $\mathbf{x}^*$.

\section{Experiments}
\begin{figure}[H]
   \centering
   \includegraphics[width=0.8\linewidth]{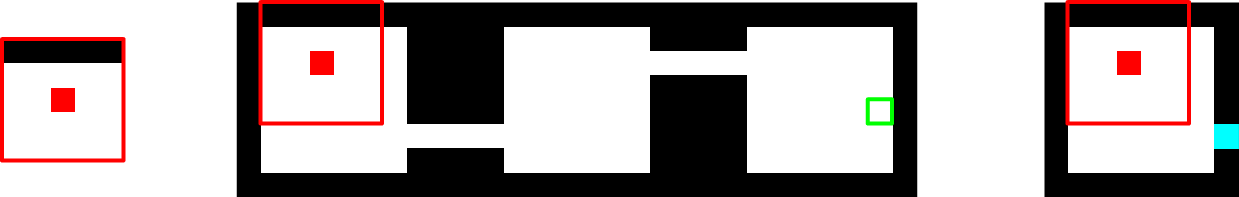}
   \caption{\textbf{Left)} Agent's egocentric observation. \textbf{Middle)} Full state/default form of PI, with the goal highlighted in green for illustration purposes. \textbf{Right)} Alternative PI, consisting of the entirety of the current room, with the correct corridor (blue) given as a sub-goal.}
   \label{fig:env}
\end{figure}

To compare different methods for leveraging PI under the RL setting, we designed a simple partially-observed environment (depicted in Figure \ref{fig:env}), where $\mathbf{x}$ is an observation of the environment, and $\mathbf{x^*}$ is a privileged feature. We initially experimented with A3C \citep{mnih2016asynchronous}, PPO \citep{schulman2017proximal} and DRQN\footnote{In addition to the original settings, we used Polyak updates for the target network and additional ``burn-in'' of the recurrent layer's hidden state \citep{kapturowski2018recurrent} when training the network.} \citep{hausknecht2015deep}, and use the latter as our baseline as it outperformed the other methods. 

\subsection{Environment}
Our environment is a gridworld consisting of three $6 \times 6$ rooms linked together. The agent (red square), which is initialised in a random position in the left or centre rooms, must navigate to the goal within 100 timesteps, receiving a reward of $-0.1$ for each action and $+10$ for reaching the goal. By default, we utilise the agent's observation, a $5 \times 5$ egocentric view, as $\mathbf{x}$, and the full state (FS) of the environment as $\mathbf{x}^*$. We also experiment with an alternative form of $\mathbf{x}^*$, in which the correct corridor to traverse---a sub-goal (SG)---is coloured blue.

\subsection{Agents}

\begin{figure}[H]
  \centering
  \includegraphics[width=0.85\linewidth]{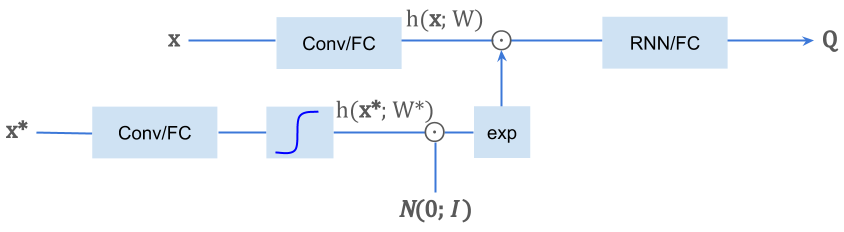}
  \caption{The architecture for a $Q$-learning-based PI-D agent. The regular features $\mathbf{x}$ and privileged features $\mathbf{x}^*$ are first processed by separate layers. During training, the intermediate representation of $\mathbf{x}$ is multiplied with log-normal noise conditioned on $\mathbf{x}^*$, while during testing, no noise is added.}
  \label{fig:architecture}
\end{figure}

The standard architecture for our DRQN-based agents consists of two convolutional layers, a GRU \citep{cho2014learning}, and two fully-connected layers, with ReLU nonlinearities. In our \pid (PI-D) agent, we use two separate convolutional sub-networks for $\mathbf{x}$ and $\mathbf{x}^*$, with multiplicative dropout applied to the output of the latter before the features of the subnetworks are concatenated (Figure \ref{fig:architecture}).

We also implemented several baselines:
\begin{itemize}
\item \textit{DRQN:} Standard agent (no PI).
\item \textit{Auxiliary task (AUX):} An agent with deconvolutional layers on top of the GRU, which is additionally trained to reconstruct $\mathbf{x}^*$ at every timestep.
\item \textit{Distillation (DIS):} A student agent that only receives $\mathbf{x}$ as input, but has an additional loss to mimic the outputs of a pre-trained teacher network that receives $\mathbf{x}^*$ as input.
\item \textit{Naive Dropout (ND):} An agent with the PI-D architecture, where the multiplicative dropout noise is manually annealed from 0 to 1 over 3000 training episodes.
\item \textit{Information Dropout (I-D):} An agent that uses information dropout (no PI).
\end{itemize}

All models are trained for $10^4$ episodes, using $\epsilon$-greedy exploration. Test performance is calculated as the reward of the greedy policy, averaged over episodes starting from all possible initial positions (in the left and centre rooms). In the results we suffix all methods with [$\mathbf{x},\mathbf{x}^*$]. For example, PI-D[5x5,FS] indicates a \pid agent that uses the egocentric view as $\mathbf{x}$ and full state as $\mathbf{x}^*$.

\subsection{Results}

\begin{figure}[H]
  \centering
  \begin{subfigure}[b]{0.49\textwidth}
    \includegraphics[width=\textwidth]{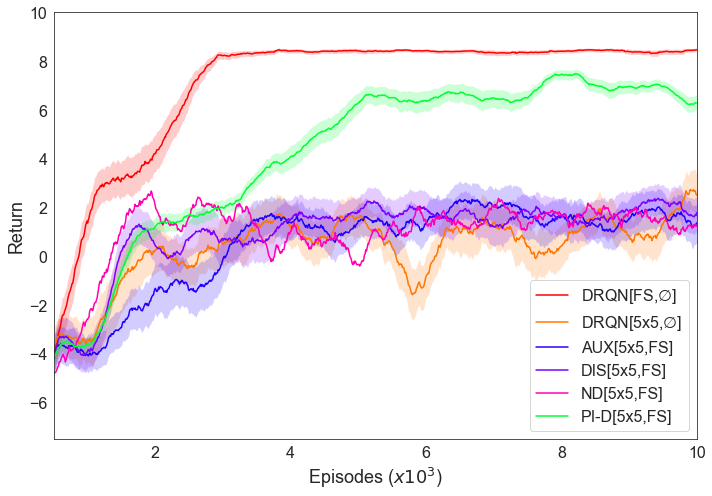}
    \caption{Comparison of baselines.}
    \label{fig:full_state}
  \end{subfigure}
  \begin{subfigure}[b]{0.49\textwidth}
    \includegraphics[width=\textwidth]{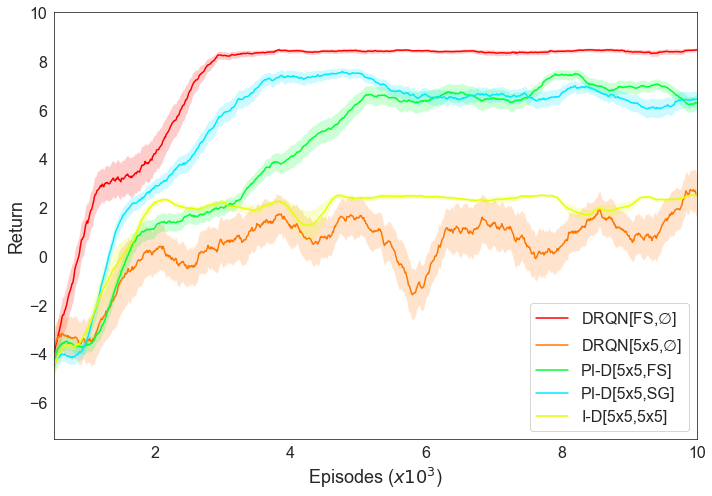}
    \caption{Comparison of forms of PI.}
    \label{fig:hrg}
  \end{subfigure}
  \caption{Average test performance of agents $\pm$ 1 standard error, calculated over 5 random seeds. $[\cdot,\cdot]$ indicates $\mathbf{x}$ and $\mathbf{x}^*$, respectively. Our method, with either full state PI (PI-D[5x5,FS]; green curve) or sub-goal PI (PI-D[5x5,SG]; cyan curve), surpasses the baselines and achieves the closest performance to that of the \textit{oracle} (DRQN[FS,$\varnothing$]; red curve).}
  \label{fig:results}
\end{figure}

The results for all the PI-based algorithms, using their best hyperparameters, are shown in the Figure \ref{fig:full_state}. The performance of the DRQN agent with full state view as input (DRQN[FS,$\varnothing$]) is plotted as an \textit{oracle} upper-bound. We also show the DRQN[5x5,$\varnothing$] as the simplest baseline. We see that our \pid agent outperforms all methods, getting closest to the performance of the oracle agent. Under detailed inspection, we observed that only our \pid agent and the oracle agent were able to consistently solve the task when initialised from any position in the environment. Other baselines fail at solving the task from the leftmost room which is the farthest from the goal, and hence show sub-optimal performance. The Naive Dropout agent (ND[5x5,FS]) in particular has poor performance across different initial positions and random seeds, highlighting the benefit of the principled \pid approach.

Figure \ref{fig:hrg} shows that \pid performs equally well, and even converges faster, with subgoals provided as PI (PI-D[5x5,SG]), demonstrating its generality. We also include an I-D[5x5,5x5] agent, which corresponds to standard information dropout (no PI). Given that its performance is the same as the DRQN[5x5,$\varnothing$] agent, this indicates that our agents benefit from PI, and not simply regularisation from dropout. In Supplementary Figure \ref{fig:beta}, we show the sensitivity of \pid to the value of $\beta$. With FS as PI, \pid is quite sensitive, but less so with SG as PI.

\subsection{Analysis}

We set out to understand the effect of \pid on our agent's learned representations, with the hypothesis that the information of the agent's position might be more easily decodable from the \pid agent, in comparison to baselines. To test this, we generated a dataset of GRU activations (which implicitly form a belief state in partially-observable environments) by collecting these from multiple rollouts for each agent, using a greedy policy, from all initial positions. We then trained linear classifiers to predict the position of the agent based on these activations.\footnote{When training the classifiers, the loss was weighted by the inverse class frequency to account for the difference in distributions of the encountered positions.} Confusion matrices for the results are shown in Supplementary Figure \ref{fig:confusion}.

The classifier for the auxiliary task agent (AUX[5x5,$\varnothing$]) had the highest accuracy of 94.8\%, which is perhaps expected given that it has the explicit auxiliary task of reconstructing the full state---and hence the agent's location. Surprisingly, the classifier for the \pid agent (PI-D[5x5,FS]) did not have a significantly higher accuracy than that of the baseline DRQN (DRQN[5x5,$\varnothing$]) classifer (88.8\% vs. at 90.1\%).
However, that solving the task successfully may not necessitate linearly decodable representations of the agent's position, and further analysis is needed to better understand the role of \pid in our trained agents. We are also interested in understanding how \pid might improve exploration, and how it compares with or can be combined with other methods that use noise to aid exploration \citep{fortunato2017noisy,plappert2017parameter}.

\section{Discussion}
In this work, we investigated the use of \pid in the context of reinforcement learning, which enables augmenting the inputs of any RL algorithms with privileged information. In a simple partially-observable environment, we demonstrated improved performance using \pid, in comparison with other methods that use PI, including distillation and auxiliary prediction tasks. We further showed the generality of \pid by utilising two different types of PI---the full state of the environment and sub-goals. An ablation against standard information dropout confirms that the use of PI is indeed beneficial and responsible for improved performance. An area for future research is to apply \pid in more challenging domains and to better understand the contributions of \pid in learning better representations or improving exploration.

\bibliography{refs}
\bibliographystyle{iclr2020_conference}

\section*{Supplementary}

\subsection*{Results}

\begin{figure}[H]
  \centering
  \begin{subfigure}[b]{0.49\textwidth}
    \includegraphics[width=\textwidth]{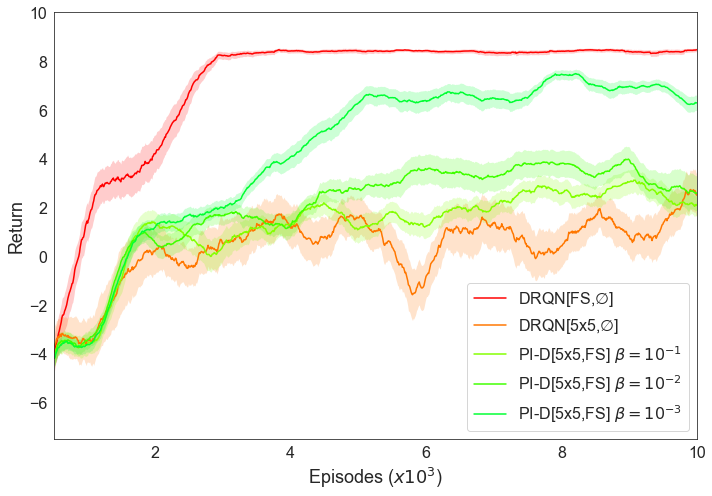}
    \caption{Full state PI.}
  \end{subfigure}
  \begin{subfigure}[b]{0.49\textwidth}
    \includegraphics[width=\textwidth]{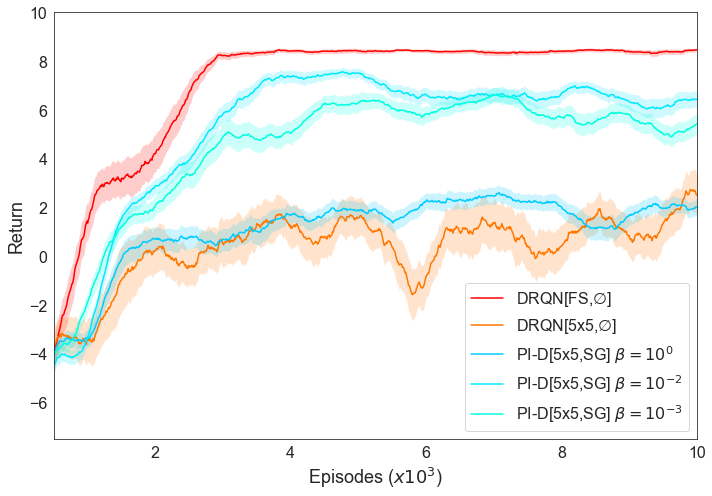}
    \caption{HRG PI.}
  \end{subfigure}
  \caption{Average test performance of agents $\pm$ 1 standard error, calculated over 5 random seeds. $[\cdot,\cdot]$ indicates $\mathbf{x}$ and $\mathbf{x}^*$, respectively.}
  \label{fig:beta}
\end{figure}

\subsection*{Analysis}

\begin{figure}[H]
  \centering
  \includegraphics[width=1\linewidth]{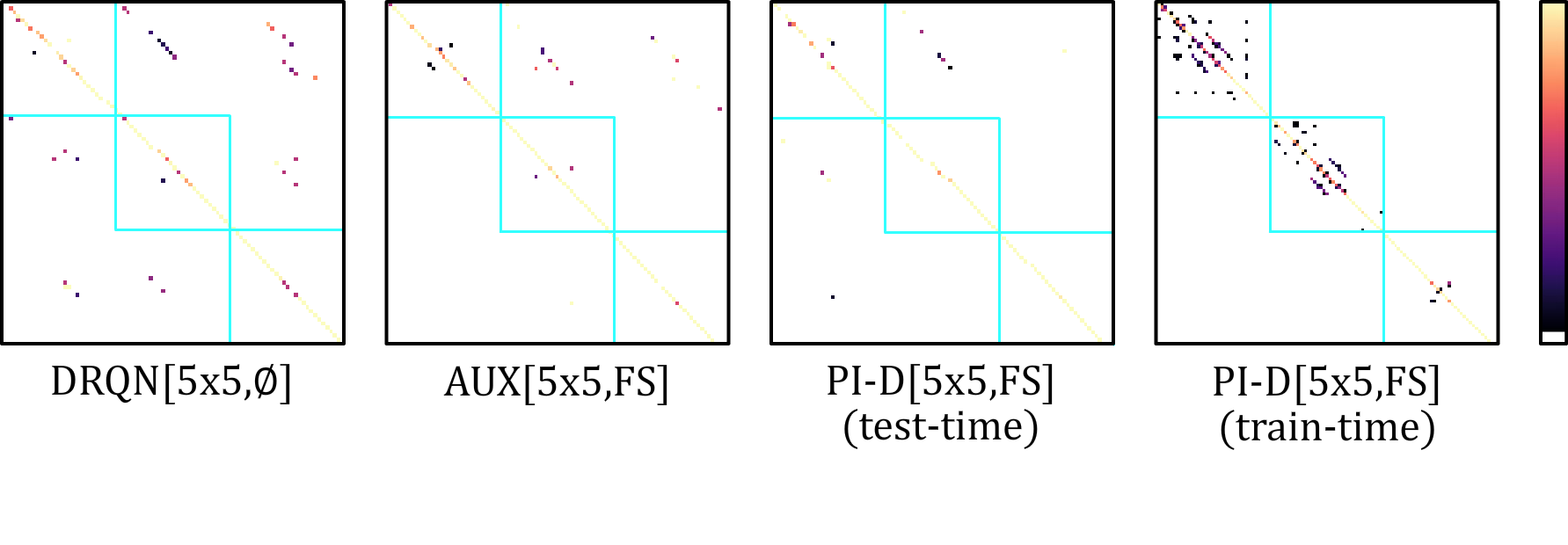}
  \caption{Real location vs. predicted location, from small (black) to large (yellow) probability mass; white indicates zero mass (to highlight low probability predictions). Room boundaries are shown in cyan. All agents without PI make errors outside of the correct room, but the \pid agent given $\mathbf{x}^*$ only makes localisation errors within the correct room.}
  \label{fig:confusion}
\end{figure}

Supplementary Figure \ref{fig:confusion}) shows confusion matrices for the linear classifiers trained on belief states from three different agents: DRQN[5x5,$\varnothing$], AUX[5x5,$\varnothing$], and PI-D[5x5,FS]. For the \pid agent, we include results from both its operation during test-time (does not receive $\mathbf{x}^*$ as input) and train-time (receives $\mathbf{x}^*$ as input). Intriguingly, during standard train-time operation the \pid agent has the highest uncertainty over the agent's location, as compared to all other settings, but it always correctly identifies the correct room, unlike in all other settings.

\end{document}